%% file: main.tex
\documentclass[11pt]{article}
\usepackage[letterpaper,margin=1in]{geometry}
\usepackage[T1]{fontenc}
\usepackage{times}
\usepackage{amsmath,amssymb,booktabs,graphicx,xcolor,multirow}
\usepackage{natbib}
\usepackage[colorlinks=true,citecolor=blue!45!black,linkcolor=blue!45!black,urlcolor=blue!45!black]{hyperref}
\usepackage{pgfplots}
\pgfplotsset{compat=1.18}
\usepgfplotslibrary{groupplots}
\usetikzlibrary{arrows.meta,fit,calc,positioning}
\usepackage{enumitem}
\setlist{leftmargin=*,itemsep=1pt,topsep=3pt}
\usepackage[font=small,labelfont=bf]{caption}
\definecolor{kcol}{HTML}{0072B2}
\definecolor{vcol}{HTML}{E69F00}
\definecolor{qcol}{HTML}{CC79A7}
\definecolor{ocol}{HTML}{009E73}

\newcommand{\F}{\mathcal{F}}
\newcommand{\Sstar}{\mathcal{S}^\star}
\newcommand{\drescue}{\Delta_{\mathrm{rescue}}}

\title{Cross-Loop Query Handoff: How Later Loops Redirect\\
Two-Hop Retrieval in the Ouro Looped Language Models}
\author{Ziyu Mu\\
{\small Faculty of Science}\\
{\small The University of British Columbia}\\
{\small\texttt{yuzzzi06@student.ubc.ca}}
\and
Zehua Wang\\
{\small Department of Electrical and Computer Engineering}\\
{\small The University of British Columbia}\\
{\small\texttt{zwang@ece.ubc.ca}}}
\date{October 23, 2026}

\usepackage{fontawesome5}
\usepackage{hyperref}

\begin{document}
\maketitle

\begin{abstract}
How do later loops in a pretrained looped language model repair two-hop retrieval? We study Ouro-1.4B
and Ouro-2.6B by transplanting attention queries between loops at the same layer, head and final prompt
position. On synthetic in-context questions, injecting the layer-indexed collection of loop-3 queries
throughout loop~1 rescues 90\% and 97\% of questions answered incorrectly at loop~1 and correctly at
loop~3, versus 24\% and 21\% for shuffled donors. The rescue also holds under greedy generation in
Ouro-2.6B; in Ouro-1.4B it is limited to four-candidate ranking. Crossed donor--recipient swaps favor
retrieval through the donor's bridge entity over copying its answer, although a substantial positional
component remains in Ouro-2.6B. Late keys and values rescue far fewer targets. Freezing queries at their
loop-1 values also damages two-hop accuracy more than a matched question that supplies the bridge.
On natural questions, rescue is smaller and replicates only in Ouro-2.6B, and crossed swaps do not
establish bridge redirection. These results identify a transferable collection of late-loop queries
that supports bridge-conditioned retrieval on synthetic scenes; they do not locate where the bridge
is computed or establish a single-vector handoff.
\end{abstract}

\vspace{0.6em}
\begin{center}
\small
\href{https://github.com/Muuzzziiy/cql-handoff}{
    \faGithub\enspace Code available on GitHub
}
\end{center}
\vspace{0.6em}

\section{Introduction}
\label{sec:intro}

Language models often fail to compose two facts they know separately. On two-hop questions about facts
stored in the weights, if the bridge entity is resolved late, the remaining layers may no longer
carry out the second hop~\citep{biran2024hopping}, and evidence for latent use of the second hop is only
moderate~\citep{yang2024latent}. Patching a later-layer state back into an earlier layer repairs some of
these failures, but only if the best layer pair is chosen for each item~\citep{biran2024hopping}. In
models trained from scratch, reusing layers on a second pass improves out-of-distribution two-hop
generalization~\citep{wang2024grokked,zhang2026twohopgen}.

Looped language models build this reuse into the architecture of pretrained models.
Ouro~\citep{zhu2025ouro} applies its layer stack four times to every token and reads out a prediction
after each loop; its authors conjecture that this recurrent structure is why small looped models
manipulate synthetic knowledge better than non-looped models of the same size~\citep[Sec.~6]{zhu2025ouro}.
Pretrained looped models have mostly been analyzed through per-loop answers~\citep{olszowiec2026monitoring},
patches between runs at the same loop~\citep{wu2026sharedweights,anon2026samehead}, or interventions on
whole loops~\citep{anon2026anotherloop,wang2026jacobian}, rather than by moving a later loop's query into
an earlier loop.

Attention offers a natural way to split the problem: at the answer position, the \emph{query} determines
what to look for; at the fact positions, the \emph{keys} determine where each fact is found and the
\emph{values} what it contributes. In a looped model, each layer and head applies the same weights in
every loop, so a query computed in loop~3 can be placed into loop~1 at exactly the same layer, head and
position, with no search over layer pairs. Our intervention supplies each layer with its own late-loop
query; it does not insert a single vector once. Sequential query mechanisms have been identified in
small models trained on two-hop tasks~\citep{guo2025twohop}. Here we test whether queries in pretrained
looped models can be transferred across recurrent passes at matching coordinates, and what retrieval
behavior that transfer induces. We study Ouro-1.4B and Ouro-2.6B on synthetic in-context
two-hop questions (``What color is in Bob's box?''), each paired with matched one-hop questions. In both
models, loop~1 answers nearly every one-hop question but fails on 42\% (1.4B) and 35\% (2.6B) of the
two-hop ones.

Our contributions are as follows:
\begin{itemize}
  \item \emph{A cross-loop query handoff.} In a setting fixed before the first run, placing the late
        queries at the answer position into loop~1, with keys and values at earlier prompt positions
        left at loop~1, corrects
        most of the loop-1 failures that loop~3 corrects, far more than another example's query does
        (Section~\ref{sec:sufficiency}); holding the query at its loop-1 value removes the loop gain
        (Section~\ref{sec:necessity}).
  \item \emph{What the handoff supports.} Crossed swaps favor retrieval through the donor's bridge
        over copying its answer, with a substantial positional component in Ouro-2.6B. Late keys and
        values recover far less (Section~\ref{sec:content}).
  \item \emph{A matched control.} Against a question that names the bridge, the query freeze is
        composition-selective in both models, and more so than freezing the values of the fact line
        both questions read (Section~\ref{sec:necessity}).
  \item \emph{A boundary on natural text.} On natural two-hop questions, the Ouro-2.6B rescue
        replicates on new items but is smaller; Ouro-1.4B shows no rescue, and a crossed swap shows no
        redirection (Section~\ref{sec:natural}).
\end{itemize}
Together, these results identify a collection of late-loop queries at the answer position that can be
transferred to an earlier loop and favors bridge-conditioned retrieval on synthetic scenes. Under this
intervention, keys and values at earlier prompt positions remain at their loop-1 states, although
freezing experiments also show that native loop gains depend on updates to fact values
(Section~\ref{sec:necessity}). We use \emph{handoff} for this transfer across matching layer and head
coordinates; the experiments neither establish a single-vector bottleneck nor locate where the bridge
is computed.

Our scope is limited. All results are on Ouro, and most are on synthetic scenes. The handoff weakens
beyond the base template. On natural questions, it is smaller in Ouro-2.6B and absent in Ouro-1.4B, which
also fails our behavioral gate on two of seven other templates (Sections~\ref{sec:natural}
and~\ref{sec:scope}); the boundary thus depends on model size and format. Ouro-1.4B's loop~1 generates
no answer, so its rescue shows only in a ranking among candidates, and our two dense controls do not
isolate looping as the source of the rescue (Section~\ref{sec:sufficiency}).

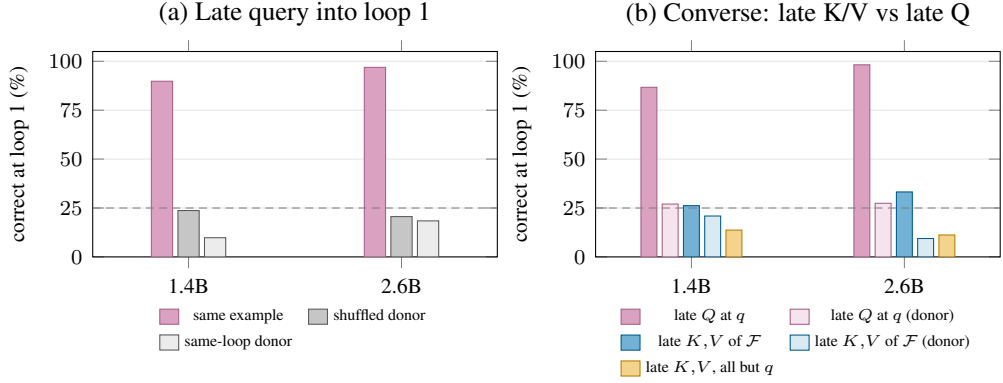
\begin{figure}[htbp]
\centering
\input{fig_rescue}
\caption{Rescue rates on two-hop targets (wrong at loop~1, right at loop~3), read at the loop-1 exit.
(a) Rescue round: the same example's loop-3 query at $q$, another target's loop-3 query (shuffled
donor), and another example's loop-1 query (same-loop donor). (b) Converse round
(Section~\ref{sec:converse}; different scenes): the late query and the late keys and values of $\F$, each
from the same example or from a shuffled donor, and the same example's late keys and values at every
position except $q$; the key--value arms leave queries unpatched. Dashed line: chance (25\%). Intervals
are in Tables~\ref{tab:rescue} and~\ref{tab:converse}; the same-loop donor and same-example late
$K,V$ of $\F$ rates are point estimates.}
\label{fig:rescue}
\end{figure}

\section{Setup}
\label{sec:setup}

\paragraph{Ouro.} Ouro~\citep{zhu2025ouro} is a family of looped decoder-only transformers. A stack of
$L$ layers ($L=24$ in Ouro-1.4B and $L=48$ in Ouro-2.6B; 16 attention heads per layer with 16 key--value
heads, head dimension 128, rotary position embeddings) is applied $T=4$ times. In loop $t$, the whole
sequence passes through the stack, and head $h$ of layer $\ell$ attends using the queries, keys and values
computed in that loop and layer from the normalized attention input $x^{(t,\ell)}_i$ at position $i$:
\begin{align*}
  Q^{(t)}_{\ell h, i} &= \mathrm{RoPE}_i\big(W^{\ell h}_Q x^{(t,\ell)}_i\big), \\
  K^{(t)}_{\ell h, j} &= \mathrm{RoPE}_j\big(W^{\ell h}_K x^{(t,\ell)}_j\big), \quad
  V^{(t)}_{\ell h, j} = W^{\ell h}_V x^{(t,\ell)}_j .
\end{align*}
Because the weights are shared across loops, a coordinate $(\ell,h,i)$ refers to the same weights in
every loop, and all differences between loops come from the inputs. The output of loop $t$ is the input
of loop $t+1$. A shared final norm and language-model head produce a next-token distribution after every
loop, which we call the \emph{loop-$t$ exit}. Ouro trains all exits jointly: the pretraining loss weights
the next-token loss at each exit by a learned exit distribution, with an entropy term that corresponds to
a uniform prior over exits~\citep[Sec.~3.3]{zhu2025ouro}. Ouro-2.6B was obtained from the 24-layer model
by duplicating its layers after the first 3 trillion training tokens~\citep[Sec.~4.3]{zhu2025ouro}, so the
two sizes are not independent replications. We read the loop-$t$ exit directly and do not use the
learned exit gate.

\paragraph{Task.} Each scene has four people, four boxes (A--D) and four colors, linked by random
bijections and written as eight fact lines (``Box D contains blue.'', ``Bob has box D.''). All box--color
lines precede all person--box lines, and the lines are shuffled within each block
(Appendix~\ref{app:prompts}). The two-hop question asks for the \emph{target} entity (a color) from the
\emph{source} entity (a person) through the \emph{bridge} entity (a box): ``Question: What color is in
Bob's box?'' followed by ``Answer:''. The matched one-hop questions in the same scene ask for the bridge
from the source (person$\to$box) and for the target from the bridge (box$\to$color). We score the four
candidates of the question's type (all single tokens) by their next-token log-probability at a loop exit;
chance is 25\%. Table~\ref{tab:notation} lists the notation. Let $a_{j,i}(t)\in\{0,1\}$ indicate
whether question type $j\in\{2,1a,1b\}$ is answered correctly in scene $i$ at loop $t$, and let
$A_j(t)=n^{-1}\sum_i a_{j,i}(t)$. The composition-specific loop gain uses scene-level changes,
$\delta_{j,i}=a_{j,i}(3)-a_{j,i}(1)$:
\[
G_{\mathrm{comp}}=\frac{1}{n}\sum_{i=1}^n
\left[\delta_{2,i}-\max\{\delta_{1a,i},\delta_{1b,i}\}\right].
\]
Thus, the larger one-hop gain is subtracted within each scene before averaging; taking the maximum of
the two aggregate one-hop gains would define a different statistic.

\begin{table}[htbp]
\centering
\caption{Notation.}
\label{tab:notation}
\small\setlength{\tabcolsep}{4pt}
\begin{tabular}{@{}lp{0.6\textwidth}@{}}
\toprule
$\F$ & the two fact lines the question uses (person--box, box--color) \\
$q$ & the final prompt position (``Answer:''), where the answer is read out \\
$\Sstar$ & scenes whose two one-hop questions are right at every loop \\
\bottomrule
\end{tabular}
\end{table}

\paragraph{Interventions.} Each intervention replaces one attention component at a fixed coordinate with
its value from another loop or another prompt and lets the rest of the forward pass run as usual; queries
and keys are replaced after RoPE.
\emph{Freezes} (dependence on native updates) hold one component at its own loop-1 value in loops 2--4, in every layer and
head: the query at $q$ ($I_Q$), or the keys ($I_K$) or values ($I_V$) at every position in $\F$.
Matched-size controls freeze the same component outside the question and outside $\F$: the keys or values
of the two unused fact lines ($C_K$, $C_V$), or the query at one other position ($C_Q$). The
\emph{retained loop gain} is $\rho_I = [A^I_2(3) - A_2(1)] / [A_2(3) - A_2(1)]$ (loop~1 is never
modified), and the \emph{selective causal damage} $\mathrm{SCD}(I)$ is the two-hop accuracy loss at
loop~3 minus the larger of the two one-hop losses (the \emph{one-hop loss}). $\rho$ measures how much of
the loop gain survives a freeze, and SCD how selectively the freeze hurts composition; we report both
because they can disagree.
The \emph{rescue} (sufficiency under query transplantation) replaces the query at $q$, in loop $r$ and in every layer and head, with
the same example's query from a later loop $s$ (the \emph{late query}):
$Q^{(r)}_{\ell h,q}\leftarrow Q^{(s)}_{\ell h,q}$ for every $(\ell,h)$.
The donor queries come from an unmodified forward pass. The recipient receives a separate cached query
at each layer and head; its residual stream and subsequent computations otherwise evolve under the
intervention. Since $q$ is the last prompt position and attention is causal, every query, key and value
at positions before $q$ keeps its native loop-$r$ value. Keys and values at $q$ itself are not patched
and may change in later layers as a consequence of earlier query patches. Thus, ``late query'' denotes
a layer-indexed collection, not a single state inserted once. Appendix~\ref{app:intervention} gives the
intervention sequence. Targets are two-hop
questions answered wrongly at loop $r$ and correctly at loop $s$. The rescue rate is the share of targets
answered correctly at the loop-$r$ exit after the patch, and $\drescue$ is this rate minus the rate for a
shuffled donor (another target's loop-$s$ query). Synthetic rescue uses donors with the same prompt
token length; natural-QA donor selection and rotary position alignment are specified in
Section~\ref{sec:natural}. The primary setting
($r=1$, $s=3$, the final position $q$, all layers and heads) was fixed before the first rescue round and
never changed; other loops and layer ranges serve as robustness checks. Rescue rates are conditional
on this target set, not accuracies over all questions or net improvements on the full set. They do not
measure whether patching would damage questions already answered correctly at loop~$r$.
\emph{Swaps} place a component from a donor prompt into a recipient prompt. Unless stated otherwise, we
read a patched model at the exit of the patched loop, and we state the donor rule for each
result~\citep{anon2026samehead}.

\paragraph{Protocol.} Each experiment had a design specifying its inputs, intervention sites, metrics,
primary tests and stopping rules, committed before that experiment was run. Follow-up experiments could
therefore be designed after earlier results were available; analyses described as post hoc were not
pre-specified primary tests. On synthetic
scenes, identity patches reproduce the unpatched outputs bitwise, and an independent script recomputed
every primary statistic. Each experiment uses newly sampled scenes: 400 per model for the freezes (each
with one two-hop and two one-hop questions) and, unless stated otherwise, 600 elsewhere. Intervals are
95\% paired bootstrap percentile intervals over scenes (over items for natural QA; 20,000 resamples),
Bonferroni-corrected within each experiment's primary tests where stated (``Bonferroni $k$'' marks a
correction over $k$ tests). Differences between rates are given in percentage points (\emph{points}).

\section{In Ouro, the loop gain is specific to two-hop composition}
\label{sec:behavior}

Before analyzing a model's mechanism, we require its loop gain to be specific to composition (the
\emph{behavioral gate}): a two-hop gain on $\Sstar$ whose 95\% interval lies above zero, and
$G_{\mathrm{comp}}>0$. On the 400 freeze scenes, two-hop accuracy rises from 58.3\% at loop~1 to
93.8\% at loop~3 in Ouro-1.4B and from 65.3\% to 99.5\% in Ouro-2.6B, gains of 35.5 points
$[30.5, 40.5]$ and 34.2 points $[29.5, 39.0]$. The one-hop questions are near ceiling from loop~1 onward
(95.5--99.5\%), with mean gains of at most 3.5 and 0.75 points (Table~\ref{tab:native}). On $\Sstar$
($n=364$ and 393), the two-hop gain is 33.8 $[28.9, 39.0]$ and 33.3 $[28.8, 38.2]$ points, and on the
full set $G_{\mathrm{comp}}$ is 28.7 $[23.5, 34.0]$ and 33.0 $[28.5, 37.8]$: the loop gain comes mostly
from composition rather than from better one-hop lookup. Both Ouro models therefore pass the gate.

\section{Late queries rescue early-loop errors}
\label{sec:sufficiency}

\begin{table}[htbp]
\centering
\caption{Query rescue at $q$, all layers and heads (600 two-hop scenes per model). Targets are wrong at
loop $r$ and right at loop $s$; rates are the share of targets correct at the loop-$r$ exit after the
patch. Same: the same example's loop-$s$ query. Shuffled: another target's loop-$s$ query. 95\% paired
bootstrap intervals.}
\label{tab:rescue}
\small\setlength{\tabcolsep}{4pt}
\begin{tabular}{lcrrrc}
\toprule
Model & $s\to r$ & $n$ & Same & Shuffled & $\drescue$ (pts) \\
\midrule
1.4B & $3\to1$ & 245 & 89.8\% & 23.7\% & \textbf{66.1} [60.0, 72.2] \\
1.4B & $3\to2$ &  91 & 100\%  & 19.8\% & 80.2 [71.4, 87.9] \\
2.6B & $3\to1$ & 223 & 96.9\% & 20.6\% & \textbf{76.2} [70.4, 81.6] \\
2.6B & $3\to2$ &  12 & 100\%  & 25.0\% & 75.0 [50.0, 100.0] \\
\bottomrule
\end{tabular}
\end{table}

Replacing only the final-position query of loop~1 with the same example's loop-3 query, while queries, keys and values at
all earlier prompt positions keep their loop-1 values, rescues 89.8\% (Ouro-1.4B) and
96.9\% (Ouro-2.6B) of the targets (Table~\ref{tab:rescue}; Figure~\ref{fig:rescue}). Another target's
loop-3 query recovers 23.7\% and 20.6\%, near chance, so $\drescue$ is 66.1 and 76.2 points; both exceed
zero with Bonferroni correction over the four model-by-loop cells (1.4B $[58.4, 73.9]$, 2.6B
$[69.1, 83.0]$). Patched into loop~2, the same query rescues every target. Another example's loop-1
query recovers 9.8\% and 18.4\%, and a shuffled donor's late query does not carry the donor's answer:
when the two answers differ, the prediction matches the donor's answer in 25.4\% and 24.8\% of cases, at
chance.

The rescued targets lie near the loop-1 decision boundary (median margin $-0.47$ and $-0.62$). After the
patch, the median margin is $+0.30$ in Ouro-1.4B, a small push across the boundary, and $+2.03$ in
Ouro-2.6B. A replication on 300 new scenes rescues 98.9\% of 91 Ouro-2.6B targets against 26.4\% for a
shuffled donor ($\drescue = 72.5$ $[63.7, 81.3]$), and three other Ouro-1.4B experiments on new scenes
rescue 81.7\%, 86.7\% and 81.8\% of their targets. Scored by greedy generation instead (600 new scenes),
the Ouro-2.6B rescue holds: the late query makes loop~1 generate the correct color for 89.6\% of 202
targets, a shuffled donor's for 20.3\% ($+69.3$ points $[61.9, 76.7]$).
Ouro-1.4B's loop~1 generates no color at all, with or without the patch, so in this model the rescue
changes which of the four candidates ranks first, not what loop~1 says.

\paragraph{Looping versus depth.} Unlike back-patching~\citep{biran2024hopping}, where the layer pair is
chosen for each item, one intervention specification fixed in advance repairs most targets here. In two dense controls, a
late query from deeper layers barely rescues: 1.7 points $[-2.5, 5.9]$ in
LayerSkip-Llama3-8B~\citep{elhoushi2024layerskip}, which is trained to answer at intermediate layers (a
query from 16 layers deeper, patched into every layer below exit~16), and at most 8.9 $[3.9, 15.1]$ in
Qwen3-1.7B-Base~\citep{yang2025qwen3} (patched at one layer; the interval extends above the 10-point bound
we set in advance). Neither control isolates looping: the dense depth offsets are at most 16 layers,
against 48 in Ouro-1.4B.

\section{Crossed query swaps favor bridge-conditioned retrieval}
\label{sec:content}

If the late query carried the answer, the model should output the donor's color; if the resolved
bridge (``look up box X''), the recipient's color for the donor's box; if a position, the recipient's
color at the line position of the donor's answer. Following the crossed design
that \citet{kale2026handoff} apply to the residual stream of non-looped models, donor and recipient ask
about the same person, who holds a different box in each, so these three predictions and the
recipient's own answer are four different colors on every item.

\begin{table}[htbp]
\centering
\caption{Two-hop query swaps (600 new scenes per model): the donor's loop-3 query at $q$, all layers and
heads, patched into the recipient at loop $r$. Shares of predictions (\%) equal to each of four
pairwise-different colors: the recipient's answer (Rec.), the recipient's color for the donor's bridge box
(Bridge), the donor's answer (Ans.), and the recipient's color at the line position of the donor's answer
(Line); chance 25\%. 95\% intervals for the bridge share.}
\label{tab:swap}
\footnotesize\setlength{\tabcolsep}{3pt}
\begin{tabular}{lcrrrr}
\toprule
Model, patch & $r$ & Rec. & \textbf{Bridge} & Ans. & Line \\
\midrule
1.4B, none & 1 & 56.3 & 10.5 & 15.3 & 17.8 \\
1.4B, swap & 1 & 15.5 & \textbf{45.0} [41.0, 49.0] & 13.3 & 26.2 \\
1.4B, swap & 3 & 14.7 & \textbf{49.2} [45.2, 53.2] & 9.0 & 27.2 \\
\midrule
2.6B, none & 1 & 68.0 & 10.3 & 11.5 & 10.2 \\
2.6B, swap & 1 & 9.0  & \textbf{48.0} [44.0, 52.0] & 8.2 & 34.8 \\
2.6B, swap & 3 & 10.8 & \textbf{45.8} [41.8, 49.8] & 7.2 & 36.2 \\
\bottomrule
\end{tabular}
\end{table}

Patched into loop~1, the donor's late query most often makes the model return the recipient's color for
the \emph{donor's} bridge box (Table~\ref{tab:swap}): 45.0\% and 48.0\%, against 13.3\% and 8.2\% for the
donor's answer. The bridge-consistent outcome is the largest category, supporting bridge-conditioned
retrieval over direct answer copying. It accounts for fewer than half of the loop-1 predictions in both
models, however, so the swap does not establish a uniformly entity-based code. Swapped at loop~3 instead, the query
has the same effect (49.2\% and 45.8\%). In Ouro-2.6B, 34.8\% and 36.2\% of predictions instead follow
the line position of the donor's answer, which is above chance (in Ouro-1.4B this share is at chance), as
with positional and pointer-based retrieval in standard
transformers~\citep{prakash2025lookbacks,gurarieh2025mixing,zhou2026ordinal}. Because line order was not
varied independently of box identity, the experiment does not fully separate entity-based and positional
routing. The positional shares are descriptive, and the query may support a mixture of retrieval
strategies. On natural questions, a
crossed swap does not show this redirection (Section~\ref{sec:natural}).

\subsection{Late keys and values recover far less than the late query}
\label{sec:converse}

\begin{table}[htbp]
\centering
\caption{Converse test (600 new scenes per model; targets wrong at loop~1 and right at loop~3, $n=263$
and 223). Each row patches the same example's loop-3 state into loop~1, except the shuffled-donor rows;
every $K,V$ row leaves queries unpatched. 95\% intervals where reported (Bonferroni 2 for the primary
contrast).}
\label{tab:converse}
\footnotesize\setlength{\tabcolsep}{2pt}
\begin{tabular}{lrr}
\toprule
Patched into loop 1 & 1.4B & 2.6B \\
\midrule
$Q$ at $q$ & 86.7\% [82.5, 90.5] & 98.2\% [96.4, 99.6] \\
$Q$ at $q$, shuffled donor & 27.0\% [21.7, 32.3] & 27.4\% [21.5, 33.2] \\
$K$ / $V$ / $K,V$ at $\F$ & 24.3 / 22.4 / 26.2\% & 4.9 / 22.4 / 33.2\% \\
$K,V$ at $\F$, shuffled donor & 20.9\% [16.0, 25.9] & 9.4\% [5.8, 13.5] \\
$K,V$ at all positions but $q$ & 13.7\% [9.5, 17.9] & 11.2\% [7.2, 15.2] \\
\midrule
Primary: $Q$ $-$ $K,V$ at $\F$ (pts) & 60.5 [52.9, 68.1] & 65.0 [57.8, 72.2] \\
\bottomrule
\end{tabular}
\end{table}

The converse test patches the same example's loop-3 keys and values into loop~1, either at $\F$ or at
every position except $q$, while leaving queries unpatched (Table~\ref{tab:converse}). Late keys and
values of $\F$ rescue 26.2\% and 33.2\%, against 86.7\% and 98.2\% for the late query; this pre-specified
primary contrast is 60.5 and 65.0 points. Patching late keys and values at all positions before $q$,
while leaving queries unpatched, rescues only 13.7\% and 11.2\%. Conversely, freezing this memory at
loop~1 removes much of the native gain (Table~\ref{tab:freeze}). These interventions show an asymmetry
between query and memory transplantation; they do not identify the causal path by which memory updates
contribute to the native gain. One possibility is that those updates help construct later queries, but
this remains untested. Net of a shuffled donor, the late keys and values of $\F$ add 5.3 points $[0.8, 9.9]$ in
Ouro-1.4B and 23.8 $[17.0, 30.5]$ in Ouro-2.6B (post hoc), against 59.7 and 70.9 for the late query: the
query transplantation is substantially more effective than memory transplantation in this setting,
with a smaller donor-specific rescue from the used fact lines in Ouro-2.6B.

\section{Freezing queries and fact values disrupts the loop gain}
\label{sec:necessity}

\begin{table}[htbp]
\centering
\caption{Freezes: in loops 2--4, one component is held at its loop-1 value. $\rho$ is the retained
two-hop loop gain at loop~3 (1 = no cost, 0 = the whole gain removed); SCD is the two-hop loss minus the
one-hop loss, which is the larger of the two one-hop accuracy losses (both in points). Matched-size
controls freeze the same component outside the question and outside $\F$. Top: 400 scenes per model.
Bottom: co-freezes on 600 further scenes (native gain 34.5 and 33.2 points). 95\% paired bootstrap
intervals.}
\label{tab:freeze}
\footnotesize\setlength{\tabcolsep}{2.5pt}
\resizebox{\textwidth}{!}{%
\begin{tabular}{lcccccc}
\toprule
 & \multicolumn{3}{c}{Ouro-1.4B} & \multicolumn{3}{c}{Ouro-2.6B} \\
\cmidrule(lr){2-4}\cmidrule(lr){5-7}
Held at loop 1 & $\rho$ & SCD (pts) & One-hop loss & $\rho$ & SCD (pts) & One-hop loss \\
\midrule
$Q$ at $q$ ($I_Q$)   & $-0.12$ [$-0.23$, $-0.02$] & 27.5 [22.2, 32.8] & 12.2 [8.8, 15.8] & $-0.14$ [$-0.26$, $-0.03$] & 36.0 [31.2, 40.8] & 3.0 [1.5, 4.8] \\
$K$ at $\F$ ($I_K$)  & 0.72 [0.62, 0.81]          & 5.8 [2.2, 9.2]    & 4.2 [2.5, 6.2]   & 0.54 [0.42, 0.64]          & 12.5 [9.0, 16.0]  & 3.2 [1.8, 5.0] \\
$V$ at $\F$ ($I_V$)  & 0.01 [$-0.16$, 0.15]       & 30.8 [25.8, 35.8] & 4.5 [2.5, 6.8]   & $-0.62$ [$-0.88$, $-0.41$] & 15.8 [11.5, 20.0] & 39.8 [35.0, 44.5] \\
$C_Q$ / $C_K$ / $C_V$ & 1.04 / 0.96 / 0.98        & $-1.2$ / 1.2 / 0.8 & 0 / 0 / 0       & 0.99 / 0.98 / 0.99         & 0.2 / 0.8 / 0.2   & 0 / 0 / 0 \\
\midrule
$V$ at $\F$                  & 0.03 [$-0.10$, 0.15]  & & & $-0.65$ [$-0.87$, $-0.46$] & & \\
$K,V$ at $\F$                & $-0.02$ [$-0.15$, 0.09] & & & 0.41 [0.31, 0.51]        & & \\
$K,V$ at all positions but $q$ & 0.35 [0.27, 0.42]   & & & $-0.06$ [$-0.14$, 0.02]    & & \\
\bottomrule
\end{tabular}}
\end{table}

The matched-size controls leave the gain intact ($\rho$ 0.96--1.04, SCD within 1.2 points of zero, no
one-hop loss; Table~\ref{tab:freeze}). This argues against damage explained solely by the number of
patched sites. It does not rule out incompatibility between early and late states at task-relevant
positions.

\paragraph{The query at the answer position.} Holding the query at $q$ at its loop-1 value removes the
whole gain and pushes two-hop accuracy below its own loop-1 level ($\rho_Q = -0.12$ and $-0.14$). By SCD,
the damage is selective for composition (27.5 and 36.0 points), but in Ouro-1.4B the freeze also costs
12.2 points on a one-hop question. The one-hop cost is smaller in Ouro-2.6B (3.0 points),
where the damage is more strongly concentrated on composition.

\paragraph{Keys and values of the used facts.} Reusing the loop-1 keys of $\F$ keeps part of the gain
($\rho_K = 0.72$ and $0.54$). Reusing their loop-1 values removes it ($\rho_V = 0.01$ and $-0.62$) and,
in Ouro-2.6B, also costs 39.8 points on a one-hop question. In Ouro-2.6B, much of this damage is
consistent with a mismatch between loop-1 values and later-loop keys: holding both the keys and the values
of $\F$ at loop~1 gives $\rho = 0.41$, against $-0.65$ for the values alone in that run (difference 1.06
$[0.85, 1.30]$, Bonferroni); in Ouro-1.4B, also holding the keys changes nothing ($-0.05$
$[-0.17, 0.07]$).

\paragraph{A matched control.} The one-hop questions differ from the two-hop question in length and
wording. A pre-specified control (400 new scenes per model) therefore pairs each two-hop question $T$
with a question $G$ that names the bridge (``What color is in box D which Bob has?'') and has the same
length, answer, candidates and fact lines. This controls these features while supplying the bridge
explicitly; the question wording also changes. We
define $M$ as $T$'s accuracy loss at loop~3 minus $G$'s. Holding the query at $q$ at its loop-1 value
costs $T$ 38.2 and 35.0 points and $G$ 21.2 and 4.8 points, so $M = 17.0$ $[11.5, 22.8]$ and $30.3$
$[25.0, 35.5]$ (Bonferroni 2): the query state is composition-selective in both models, although in
Ouro-1.4B it also partly serves generic retrieval. In Ouro-2.6B, $G$ is near ceiling from loop~1 onward
(96.5\%), so there $M$ mostly reflects the fact that only $T$ has a loop gain.
A second pre-specified run, on new scenes, freezes each used line separately. For the box$\to$color line,
which both questions read, freezing the values costs $T$ and $G$ about equally ($M = -1.2$ and $4.8$) and
keeps most of the gain ($\rho = 0.69$ and $0.64$); this freeze is less selective than the query freeze by
14.3 $[8.8, 19.8]$ and 22.5 $[16.8, 28.3]$ points (Bonferroni 2). Freezing the values of the
person$\to$box line, which only $T$ reads, removes most or all of $T$'s gain ($\rho = 0.26$ and
$-0.85$); this design cannot tell whether that reflects composition or only the fact that $G$ does not
read the line. On matched material, the clearest composition signal thus comes from the query, while
later loops also rewrite part of the memory.

Under these freezes, preserving the native loop gain depends on updates to both queries and fact
values, with interactions between keys and values. Conversely, the query rescue tests whether late
queries can recover answers while the memory at earlier prompt positions remains at loop~1.

\section{Natural two-hop questions}
\label{sec:natural}

We test the primary setting on two-hop questions from 2WikiMultiHopQA~\citep{ho2020twowiki} and
MuSiQue~\citep{trivedi2022musique}, given with their gold passages in Ouro's chat format, and score four
candidate answer strings by length-normalized log-likelihood (Appendix~\ref{app:prompts}).

For the initial rescue and fresh replication, shuffled donors are seeded permutations of the target
set with no item assigned to itself, without matching prompt lengths. Before patching, each cached
post-RoPE donor query is re-rotated by $\Delta p=L_{\mathrm{recipient}}-L_{\mathrm{donor}}$, where $L$ is
the prompt token length, to align the donor's final prompt position with the recipient's at each
layer and head. Same-item patches have zero offset. Last-quarter controls also allow different prompt
lengths and use the same alignment; natural-QA crossed swaps use their stored donor--recipient
pairings with the same rotary alignment.

\begin{table}[htbp]
\centering
\caption{Natural-QA query rescue at loop~1 using loop-3 queries. Initial set: 1,105 items from
2WikiMultiHopQA and MuSiQue; replication: 646 unused 2WikiMultiHopQA items. $n$ counts targets wrong at
loop~1 and right at loop~3. Rescue and shuffled-donor rates are percentages; $\drescue$ is in points.
Intervals are 95\%; both 2.6B replication intervals (all layers and last quarter) are
Bonferroni-corrected over two primary tests.
Last-quarter patches are follow-up analyses.}
\label{tab:natural}
\small\setlength{\tabcolsep}{4pt}
\begin{tabular}{lllrrrr}
\toprule
Model & Set & Layers & $n$ & Rescue & Shuffled & $\drescue$ [95\% CI] \\
\midrule
1.4B & Initial & All & 176 & 7.4 & 6.3 & 1.1 [$-1.1$, 3.4] \\
2.6B & Initial & All & 134 & 33.6 & 19.4 & 14.2 [7.5, 20.9] \\
2.6B & Replication & All & 110 & 31.8 & 23.6 & 8.2 [1.8, 15.5] \\
2.6B & Initial & Last quarter & 134 & 62.7 & 32.8 & 29.9 [20.9, 38.8] \\
2.6B & Replication & Last quarter & 110 & 63.6 & 38.2 & 25.5 [16.4, 35.5] \\
\bottomrule
\end{tabular}
\end{table}

\paragraph{Ouro-2.6B: a smaller rescue that replicates.} On 1,105 items, the late query rescues 33.6\% of
134 targets, and a shuffled donor's 19.4\% ($\drescue = 14.2$ points $[7.5, 20.9]$). A
pre-specified replication on 646 unused items (all from 2WikiMultiHopQA, as no unused MuSiQue items
remained) gives 31.8\% against 23.6\% of 110 targets ($+8.2$ $[1.8, 15.5]$, Bonferroni 2), smaller
than the first estimate. Patching only the last quarter of the layers (all heads) rescues more:
62.7\% against 32.8\% for a shuffled donor ($+29.9$ $[20.9, 38.8]$; a follow-up), and this
result replicates on the new items (63.6\% against 38.2\%, $+25.5$ $[16.4, 35.5]$). Because another
item's late-layer query alone recovers roughly a third of the targets, the absolute rescue rate
includes a substantial component not specific to the donor item.

\paragraph{Ouro-1.4B: a loop gain without a handoff.} On the same 1,105 items, Ouro-1.4B gains 15.0
points $[12.8, 17.3]$ from looping (78.3\% to 93.3\%), but its late query does not carry the gain into
loop~1: 7.4\% against 6.3\% of 176 targets ($+1.1$ $[-1.1, 3.4]$). As on two of the template variants
(Section~\ref{sec:scope}), the smaller model's handoff depends more on the input; we report this as a
boundary condition and do not explain it.

\paragraph{No redirection in a crossed swap.} To repeat the crossed swap of Section~\ref{sec:content} on
natural text, we add one passage to the recipient's prompt: the donor's second-hop passage, with its
answer replaced by another answer of the same type. Following the donor's bridge, copying the donor's
answer and giving the recipient's own answer then correspond to different candidates. In a pre-specified
test on the 711 pairs in which Ouro-2.6B answers both items correctly at loop~3, the donor's late query
makes the model follow the donor's bridge on 27.7\% of pairs, and an unrelated item's late query on
26.2\% ($+1.5$ $[-0.1, 3.2]$, Bonferroni 2; not supported); patching only the last quarter of the layers
gives the same result ($+1.8$ $[0.0, 3.7]$). Without any patch, the edited passage alone draws this
choice on 25.7\% of pairs, and the donor's answer is almost never chosen (at most 7\% in any arm). On
these items, we therefore cannot establish that the late query carries the resolved intermediate.
The positive rescue result alone does not establish the synthetic bridge-retrieval mechanism on natural
text. Nor are freezes on natural items
composition-selective against a bridge-given question (SCD within 2 points of zero).

\section{The rescue weakens beyond the base template}
\label{sec:scope}

\paragraph{Templates and formats.} On three template variants (V1 shuffles the facts and adds distractor
lines; V2 rewords the names, containers, attribute, relation and question; V3 does both), Ouro-2.6B passes
the gate and the rescue still works, though more weakly (rescue 44.0--76.5\% of targets against
96.9\%); Ouro-1.4B fails the gate on V2 and V3 ($G_{\mathrm{comp}} = -6.0$ and $-8.8$). On four formats
with shuffled facts (lockers, rooms, cars and a short story), gate and rescue pass in all eight cells,
again more weakly: the late query rescues 60--79\% of targets in Ouro-2.6B
and 52--71\% in Ouro-1.4B, where a shuffled donor gets 16--45\%.

\paragraph{Larger scenes.} With 8 people, boxes and colors, distractor lines and eight-way scoring, both
models pass the gate and the rescue holds ($\drescue$ 30.5 and 35.5 points; rescue 38--48\% of targets,
shuffled donor 7.0--12.9\%). With 16 of each, Ouro-2.6B still passes ($\drescue = 26.3$),
and Ouro-1.4B fails the gate. On the four formats with 8 and 16 entities, Ouro-2.6B passes the gate in all
eight cells and Ouro-1.4B in six: it fails the gate on the locker and story formats at 16 entities, which
leaves the rescue untested there. Wherever the gate passes, the late query beats a shuffled donor's by
16--35 points in Ouro-2.6B and by 18--29 in Ouro-1.4B, while the rescue rate falls from 31--48\% of
targets at 8 entities to 20--25\% at 16 ($k$-way scoring).

\section{Related work}
\label{sec:related}

\paragraph{Latent multi-hop reasoning and patching.} Letting lower layers attend to higher layers
improves multi-hop answers~\citep{yu2025back}, and Patchscopes repair two-hop failures by patching a
first-hop representation into a second-hop prompt~\citep{ghandeharioun2024patchscopes}. In a three-layer
toy model of in-context two-hop reasoning, a later layer's query built from the copied bridge retrieves the
target~\citep{guo2025twohop}, providing a precedent for the sequential query mechanism (see also
\citealp{ye2025implicit}). Our contribution is to test transfer across loops at matching coordinates in
pretrained models and to distinguish bridge-consistent retrieval from answer copying. Query patching
itself is an established tool: filter-head queries carry a predicate~\citep{sensharma2026filter}; our
crossed swaps instead test entity-conditioned retrieval. Bound entities are retrieved by
pointer and positional mechanisms~\citep{prakash2025lookbacks,gurarieh2025mixing,zhou2026ordinal}, and
routing heads carry multilingual two-hop reasoning in dense LMs~\citep{kim2026bridgerouting}.

\paragraph{Looped language models.} Weight-shared depth goes back to the Universal
Transformer~\citep{dehghani2019universal}, and looping has been argued to suit compositional
reasoning~\citep{saunshi2025latent,popescu2026toolcalling}. Interpretability work on looped LMs finds
fixed points with stable head behavior and heads that keep their roles across
iterations~\citep{blayney2026mechanistic,olszowiec2026monitoring}, limited evidence of decodable
intermediate results in Huginn~\citep{lu2025latentcot}, cross-loop structure in Ouro and
Huginn~\citep{wang2026jacobian}, and, in Ouro-1.4B, an intermediate of two-step arithmetic that a probe
reads but the model does not use~\citep{shalevmerin2026present}. In models trained from scratch,
bridge entities become causal and decodable over
loops~\citep{kohli2026loop,fu2026discoloop,anon2026storedfacts}, and a recurrent search circuit supports
depth generalization~\citep{notallthinking2026}.
\citet{wu2026sharedweights} swap attention patterns (and, in an appendix, queries) between runs at the
same loop in an Ouro-2.6B model fine-tuned on letter transitivity, with one cross-loop control from
loop~2 into loop~4. In the released Ouro-2.6B, \emph{Same Head, Different Calls?}~\citep{anon2026samehead}
shows that a head-output patch can lose its effect one loop later, and \emph{What Does Another Loop
Buy?}~\citep{anon2026anotherloop} shows that freezing a loop's attention pattern at its previous value
hurts GSM8K. None of these moves a later loop's query into an earlier loop, compares
query, key and value freezes at fixed coordinates, or tests two-hop rescue. Methods that compress or share
the KV cache across
loops~\citep{oneill2026lla,yang2026flashloop,kim2026residualquant,monea2026shared,conchello2026melt,ladhak2026elastic,mishra2026dynamic,huang2026loopedright}
or layers~\citep{wu2025crosslayer,lin2026fusedkv} mostly train or adapt the model for sharing. In Huginn,
queries and keys approach their final geometry before values do~\citep{shallowqueries2026}, and keys and
attention routing settle early in looped models~\citep{oneill2026lla,wan2026wise}; in Ouro, the query
still carries a needed update.

\section{Limitations}
\label{sec:discussion}

We do not show where the composition is computed: loop~1 binds each person to a box (97.5\% and 99.5\% on
person$\to$box) yet fails on 42\% and 35\% of two-hop questions. Both models come from the Ouro family,
and Ouro-2.6B was grown from the 24-layer model; we make no claim about looped models in general. Our
dense controls are only two models, with depth offsets of at most 16 layers against 48 in Ouro-1.4B, so
we show neither that looping causes the rescue nor that dense transformers lack it. The mechanism results
come from synthetic scenes, and the freezes, swaps and converse test were not rerun on other templates.
On natural questions, the handoff is smaller in Ouro-2.6B (the replication covers 2WikiMultiHopQA only),
absent in Ouro-1.4B, and not shown to redirect retrieval. For the person$\to$box values, the matched
control cannot separate composition from the fact that only the two-hop question reads that line, and in
Ouro-2.6B the bridge-given question is near ceiling. Patched and frozen states are hybrids that the model
never produces natively; our controls bound generic damage but not the distance of these states from
natural ones. Query transplantation supplies late-loop information separately at every layer and head;
it does not show that a single query vector suffices, or that the recipient independently recomputes the
bridge. The crossed swap also leaves positional routing incompletely separated from entity identity.
Rescue is measured on questions selected for a loop-1 error and a loop-3 success; it does not establish
the net benefit of patching all questions. In Ouro-1.4B, the rescue moves items just across the decision boundary of a four-candidate
ranking, and loop~1 generates no answer, with or without the patch.

\section{Conclusion}
\label{sec:conclusion}

On synthetic two-hop questions, transplanting the layer-indexed collection of Ouro's late-loop queries
at the answer position rescues most selected early-loop errors. Crossed swaps favor retrieval through
the donor's bridge over answer copying, while leaving a substantial positional component in Ouro-2.6B.
Query freezes also damage two-hop questions more than matched questions that supply the bridge.
On natural questions, a smaller rescue replicates in Ouro-2.6B, but bridge redirection is not established.
These results identify transferable queries that support synthetic two-hop retrieval. Where the bridge
is computed, how memory updates contribute, and whether a single-state handoff can suffice remain open.

\section*{Impact statement}
This paper analyzes how a pretrained looped language model answers two-hop questions. A better
understanding of where answers are read out may help to diagnose reasoning failures in such models. We
do not see societal consequences specific to this work that need to be highlighted here.

\bibliographystyle{plainnat}
\bibliography{refs}

\clearpage
\appendix

\section{Prompt formats}
\label{app:prompts}

\paragraph{Synthetic two-hop.} Raw text, without a chat template. A two-hop prompt, verbatim:
\begin{verbatim}
Use only the facts in this scenario.

Box B contains red.
Box D contains blue.
Box C contains yellow.
Box A contains green.
Bob has box D.
Carol has box C.
Alice has box B.
David has box A.

Question: What color is in Bob's box?
Answer:
\end{verbatim}
The two one-hop prompts are identical except for the question: they end instead with ``Question: Which
box does \emph{person} have?'' followed by ``Answer: box'' (candidates `` A''--`` D''), or with
``Question: What color is in box \emph{b}?'' followed by ``Answer:''. Each question is a separate prompt,
and every candidate is a single token scored at the next position.

\paragraph{Natural QA.} The prompt, with passage text abbreviated; candidate answers are scored after the
final newline:
{\footnotesize
\begin{verbatim}
<|im_start|>system
You are a helpful assistant.<|im_end|>
<|im_start|>user
Use only the following passages to answer the question. Give only the short answer.

Passages:
Title: <gold passage 1 title>
<gold passage 1 text>

Title: <gold passage 2 title>
<gold passage 2 text>

Question: <question><|im_end|>
<|im_start|>assistant
\end{verbatim}
}
The position $q$ is the last prompt token: the newline after the final \verb+<|im_start|>assistant+. The
three distractors are gold answers of other items from the same dataset with the same answer type.

\section{Query transplantation procedure}
\label{app:intervention}

The primary rescue uses a native donor run and an intervened recipient run of the same prompt. Crossed
swaps use different prompts. The mathematical intervention proceeds as follows:
\begin{enumerate}
\item Run the donor without intervention through loop $s$ and retain its post-RoPE queries
      $\{Q^{(s)}_{\ell h,q}\}_{\ell,h}$ at the final prompt position.
\item Run the recipient to loop $r$. At each layer $\ell$, replace the post-RoPE query at $q$ in
      every head $h$ with the donor query at the matching $(\ell,h)$ coordinate.
\item Compute attention and the remainder of the layer with this replacement, then continue to the
      next layer, where its corresponding donor queries are supplied. No key, value or residual-stream
      activation is directly replaced in the query-rescue arm.
\item Read the recipient's output at the loop-$r$ exit. For rescue, score only targets that were
      wrong at native loop $r$ and right at native loop $s$, using the same targets for the
      same-example and shuffled-donor arms.
\end{enumerate}
For the primary setting, $r=1$ and $s=3$. Causal attention ensures that positions before $q$ are
unaffected by these final-position query patches, so their keys and values remain native to the
recipient loop. At $q$, however, the residual stream and later-layer keys and values can respond to
previous layers' patches. The resulting forward pass is therefore a hybrid computation supplied with
late-loop queries throughout the stack, rather than a native early-loop computation initialized by
one late-loop vector. The cached donor information already required the donor's later-loop computation;
the rescue experiment is a mechanistic intervention, not a demonstrated reduction in inference cost.

\section{Native accuracy by loop}
\label{app:native}

\begin{table}[htbp]
\centering
\caption{Native forced-choice accuracy (\%) at the loop-$t$ exit. Freeze round: 400 scenes per model,
each with one two-hop and two one-hop questions. Rescue round: 600 two-hop scenes per model, screened at
loops 1--3. In Ouro-1.4B, two-hop accuracy falls from loop~3 to loop~4.}
\label{tab:native}
\small
\begin{tabular}{llrrrr}
\toprule
Model & Question & loop 1 & loop 2 & loop 3 & loop 4 \\
\midrule
1.4B & two-hop        & 58.25 & 76.25 & 93.75 & 87.50 \\
     & person$\to$box  & 97.50 & 100.00 & 100.00 & 100.00 \\
     & box$\to$color   & 95.50 & 98.00 & 99.00 & 98.25 \\
     & two-hop (rescue round) & 53.0 & 77.5 & 92.5 & -- \\
\midrule
2.6B & two-hop        & 65.25 & 97.00 & 99.50 & 99.25 \\
     & person$\to$box  & 99.50 & 100.00 & 100.00 & 100.00 \\
     & box$\to$color   & 99.25 & 100.00 & 100.00 & 99.50 \\
     & two-hop (rescue round) & 62.5 & 97.83 & 99.67 & -- \\
\bottomrule
\end{tabular}
\end{table}

In Ouro-1.4B, two-hop accuracy drops from 93.75\% at loop~3 to 87.50\% at loop~4, while both one-hop
questions stay at 98--100\%; Ouro-2.6B shows no such drop (99.50\% and 99.25\%). We did not study this
drop. Every Ouro loop gain in this paper is measured at loop~3.

\end{document}

%% file: fig_rescue.tex
\begin{tikzpicture}
\begin{groupplot}[
  group style={group size=2 by 1, horizontal sep=1.3cm},
  width=0.42\textwidth, height=4.3cm,
  ybar, ymin=0, ymax=105,
  ylabel={correct at loop 1 (\%)},
  symbolic x coords={1.4B,2.6B}, xtick=data,
  enlarge x limits=0.45,
  ytick={0,25,50,75,100},
  ymajorgrids, grid style={gray!20},
  tick label style={font=\scriptsize}, label style={font=\scriptsize},
  title style={font=\small},
  legend style={font=\tiny, draw=none, fill=none, at={(0.5,-0.2)}, anchor=north, legend columns=2, /tikz/every even column/.append style={column sep=4pt}},
  legend image code/.code={\draw[#1] (0cm,-0.1cm) rectangle (0.25cm,0.1cm);},
]
\nextgroupplot[bar width=8pt, title={(a) Late query into loop 1}]
\addplot[fill=qcol!70, draw=qcol!80!black] coordinates {(1.4B,89.8) (2.6B,96.9)};
\addplot[fill=gray!45, draw=gray!70!black] coordinates {(1.4B,23.7) (2.6B,20.6)};
\addplot[fill=gray!15, draw=gray!70!black] coordinates {(1.4B,9.8) (2.6B,18.4)};
\legend{same example, shuffled donor, same-loop donor}
\draw[densely dashed, gray] (rel axis cs:0,0.238) -- (rel axis cs:1,0.238);
\nextgroupplot[bar width=6pt, title={(b) Converse: late K/V vs late Q}]
\addplot[fill=qcol!70, draw=qcol!80!black] coordinates {(1.4B,86.7) (2.6B,98.2)};
\addplot[fill=qcol!20, draw=qcol!80!black] coordinates {(1.4B,27.0) (2.6B,27.4)};
\addplot[fill=kcol!55, draw=kcol!80!black] coordinates {(1.4B,26.2) (2.6B,33.2)};
\addplot[fill=kcol!15, draw=kcol!80!black] coordinates {(1.4B,20.9) (2.6B,9.4)};
\addplot[fill=vcol!45, draw=vcol!80!black] coordinates {(1.4B,13.7) (2.6B,11.2)};
\legend{late $Q$ at $q$, late $Q$ at $q$ (donor), {late $K{,}V$ of $\mathcal F$}, {late $K{,}V$ of $\mathcal F$ (donor)}, {late $K{,}V$, all but $q$}}
\draw[densely dashed, gray] (rel axis cs:0,0.238) -- (rel axis cs:1,0.238);
\end{groupplot}
\end{tikzpicture}